\documentclass{article}

\usepackage{PRIMEarxiv}

\usepackage[utf8]{inputenc} 
\usepackage[T1]{fontenc}    
\usepackage{hyperref}       
\usepackage{url}            
\usepackage{booktabs}       
\usepackage{amsfonts}       
\usepackage{nicefrac}       
\usepackage{microtype}      
\usepackage{lipsum}
\usepackage{fancyhdr}       
\usepackage{graphicx}       
\graphicspath{{media/}}     
\usepackage{todonotes}
\usepackage{booktabs}
\usepackage{multirow}

\usepackage[english]{babel}

\usepackage{amsmath}
\usepackage{graphicx}

\title{LiLiuM: eBay's Large Language Models for e-commerce}

\author{
  \textbf{Christian Herold} \quad \textbf{Michael Kozielski} \quad \textbf{Leonid Ekimov} \\
  \textbf{Pavel Petrushkov} \quad \textbf{Pierre-Yves Vandenbussche} \quad \textbf{Shahram Khadivi}\\
  \vspace{0.5cm} \\ eBay Inc. \\
  \texttt{\{cherold, skhadivi\}@ebay.com} \\
}

\begin{document}
\maketitle

\begin{abstract}

We introduce the LiLiuM series of large language models (LLMs): 1B, 7B, and 13B parameter models developed 100\,\% in-house to fit eBay's specific needs in the e-commerce domain.
This gives eBay full control over all aspects of the models including license, data, vocabulary, and architecture.
We expect these models to be used as a foundation for fine-tuning and instruction-tuning, eliminating dependencies to external models.

The LiLiuM LLMs have been trained on 3 trillion tokens of multilingual text from general and e-commerce domain.
They perform similar to the popular LLaMA-2 models on English natural language understanding (NLU) benchmarks.
At the same time, we outperform LLaMA-2 on non-English NLU tasks, machine translation and on e-commerce specific downstream tasks.

As part of our data mixture, we utilize the newly released RedPajama-V2 dataset for training and share our insights regarding data filtering and deduplication.
We also discuss in detail how to serialize structured data for use in autoregressive language modeling.
We provide insights on the effects of including code and parallel machine translation data in pre-training.
Furthermore, we develop our own tokenizer and model vocabulary, customized towards e-commerce.
This way, we can achieve up to 34\% speed-up in text generation on eBay-specific downstream tasks compared to LLaMA-2.

Finally, in relation to LLM pretraining, we show that checkpoint averaging can further improve over the best individual model checkpoint.

\end{abstract}


\section{Introduction}

Large Language Models (LLMs) have become the center point of natural language processing applications.
They power technologies such as OpenAI's ChatGPT \cite{DBLP:journals/corr/abs-2303-08774}, Anthropic's Claude \cite{claude3} and Google's Gemini \cite{DBLP:journals/corr/abs-2312-11805}, among many others.
At the heart of these technologies are large foundation LLMs that are trained on huge amounts of text data.

There exist foundation models that can be accessed and tuned for specific use-cases, such as the LLaMA-2 models from meta \cite{DBLP:journals/corr/abs-2307-09288}.
However, using these models poses a risk in terms of licensing, data safety and future proofing among other things.
Also, these models are very generic and mostly trained on English-centric data.
In this work, we introduce eBay's own series of LLM foundation models, which we call LiLiuM.
The LiLiuM models
\begin{itemize}
    \addtolength\itemsep{-0.67mm}
    \item have comparable performance to LLaMA-2 on English natural language understanding (NLU) benchmarks.
    \item outperform LLaMA-2 on non-English NLU benchmarks.
    \item outperform LLaMA-2 on machine translation (MT) benchmarks.
    \item outperform LLaMA-2 on e-commerce benchmarks.
    \item provide faster inference on e-commerce tasks, thanks to vocabulary customized towards the e-commerce domain.
\end{itemize}
These models are meant to eliminate dependency on third party LLMs within eBay.

The rest of this work is structured as follows.
In Section \ref{sec:lilium_framework} we describe the overall framework in which the LiLiuM models are developed, including the software, hardware, data and evaluation.
In Section \ref{sec:finding_best_setup} we describe the series of experiments we conducted in order to find the best overall setup for model training.
Finally, in Section \ref{sec:lilium_models} we describe the final 1B, 7B and 13B models we release.

\section{LiLiuM Framework}
\label{sec:lilium_framework}

In this section we describe the general framework in which we develop our models.

\subsection{Training Framework and Hardware}
\label{subsec:framework_and_hardware}

We base our training framework on Megatron-LM from NVIDIA \cite{DBLP:journals/corr/abs-1909-08053, DBLP:conf/sc/NarayananSCLPKV21} which we customize to our specific use-case.
Specifically, we add support for our tokenizer format and add support for continued training with new data.
Megatron-LM is a highly optimized training framework that allows us to use 3D parallelism in training (data parallel (DP), tensor parallel (TP), pipeline parallel (PP)) as well as distributed optimizer states \cite{rajbhandari2020zero}.
In addition, it makes use of mixed-precision training \cite{DBLP:journals/corr/abs-1710-03740}, FlashAttention-2 \cite{DBLP:journals/corr/abs-2307-08691}, and optimized data loading, among other optimizations.

Training was conducted using 100 nodes, each having 8 NVIDIA A100 80GB GPUs (a total of 800 GPUs).
The GPUs are connected via NVIDIA NVLink (intra-node) and InfiniBand (inter-node).
The hardware is part of the NVIDIA DGX cloud platform.

For the 1B model training, we utilize only data parallelism.
For the 7B and 13B models, we run a set of experiments to determine the most efficient model distribution setting for the given hardware setup (see Table \ref{tab:lilium_setup}).
We note that the optimal setting very much depends on other hyperparameters of the training, such as total number of GPUs and amount of gradient accumulation.
So our optimal setup might not be transferable to other scenarios.

\subsection{Architecture}
\label{subsec:architecture}

For the architecture of the LiLiuM models, we mostly follow existing work \cite{DBLP:journals/corr/abs-2307-09288, jiang2023mistral}.
Specifically, we adopt a decoder-only transformer architecture \cite{DBLP:conf/nips/VaswaniSPUJGKP17} with a context size of 4096 tokens, rotary position embeddings \cite{DBLP:journals/ijon/SuALPBL24} and SwiGLU activation function \cite{DBLP:journals/corr/abs-2002-05202}.
For the 1B model, we optimize the model architecture in terms of number of layers vs number of parameters per layer, see Section \ref{subsec:best_1b_arch}.
The exact model architectures are outlined in Table \ref{tab:lilium_setup}.

\subsection{Data}
\label{subsec:data}

In this section we describe the datasets utilized for training the LiLiuM models, summarized in Table \ref{tab:data_mixture}.
We also want to highlight our approach towards data curation and preprocessing. In contrast to previous work, our strategy encompasses a dual focus: embracing multilinguality and incorporating e-commerce specific data. 
For non-English languages, in this work, we focus on German, Spanish, Italian and French. 

\begin{table}[h!]
\caption{Dataset mixture for the LiLiuM models. \lq{}multilingual\rq{} in this context refers to the languages English, German, Spanish, Italian and French. \lq{}Sample Ratio\rq{} refers to the ratio with which the datasets were sampled during training.\label{tab:data_mixture}}
\centering
\def\arraystretch{1.2}
\begin{tabular}{l|l|r|r|r}
Dataset                    & Type & Language         & \# tokens & Sample Ratio \\ \midrule
e-commerce                 & Listings \& Products  & multilingual & 5,330B          & 10.0\,\%             \\
RefinedWeb                 & General web  & English & 575B          & 19.2\,\%             \\
RedPajama-V2 (newest 15 snapshots)               & General web  & multilingual & 3,100B          & 59.6\,\%             \\ 
The Stack                  & Code         & programming & 281B          & 5.0\,\%             \\ 
StackExchange              & Q\&A         & English& 3B          & 0.2\,\%             \\
peS2o                      & Academic     & English& 59B          & 3.9\,\%             \\ 
Wikipedia                  & Encyclopedic & multilingual& 9B          & 0.6\,\%             \\ 
Machine Translation & Translation   & multilingual & 42B          & 1.4\,\%
\end{tabular}
\end{table}

\subsubsection{e-commerce}
\label{subsec:e_commerce}

We source e-commerce data from two primary channels. The first source consists of user-generated e-commerce listings, spanning from the year 2018 up to January 2024. These listings feature a title, a description, and multiple name/value pairs detailing the product. Additionally, we collect user queries used to search for these listings, along with metadata such as condition, price, and listing type.

To refine this dataset, we apply several rounds of filtering. We began by excluding private listings and those from sellers opting out of data usage. We then employ heuristic filters based on the length of titles and descriptions. The final step involves deduplication of the data, based on titles and descriptions. This is achieved through fuzzy deduplication, which entails normalizing titles and descriptions before performing exact deduplication. This process yields a dataset of 7.3 billion listings consisting of 5.3 trillion tokens of text.

The second data source is our comprehensive internal product catalog, which also encompasses titles, descriptions, and name/value pairs, albeit with significantly higher quality due to its aggregated and curated nature. Moreover, for many products, we have linked reviews, ranging from a handful to several thousand, depending on the product's popularity.

The filtering process for the product data involves choosing only active products with associated active listings. For the reviews, we select those that passed the spam filters and were displayed on the site, excluding any review with low relevance scores as determined by an internal classifier. This results in a dataset of 200 million products consisting of 30 billion tokens of text.

Given the structured data type of the listings and products, we need to to serialize this data to be used in autoregressive LM training.
To maximize the versatility of the LM, we randomize the content order during serialization.
Furthermore, we experiment with several ways of indicating different listing attributes to the model, as described in Section \ref{subsec:serializing_ebay_data}.

\subsubsection{General Domain}
\label{subsubsec:data_general_domain}

For the general domain data, we first consider the public available version of RefinedWeb \cite{DBLP:journals/corr/abs-2306-01116}, which is a high quality dataset for LLM pre-training, used to train the Falcon family of LLMs  \cite{DBLP:journals/corr/abs-2311-16867}.
However, the dataset has two main disadvantages. First, it is English-language only and second, it is too small for our purposes (ca. 575 billion tokens using our tokenizer). 

Therefore, we additionally utilize the newly released RedPajama-V2 dataset \cite{together2023redpajama}, which is multilingual\footnote{We find that after filtering, around 36\,\% of the corpus is non-English} and consists of tens of trillions of tokens.
In contrast to RefinedWeb, RedPajama-V2 has not been filtered and deduplicated, but instead provides pre-computed quality signals and LSH signutures as parts of the data release. These quality signals allow us to apply a similar filtering as used in other datasets like RefinedWeb, Gopher \cite{DBLP:journals/corr/abs-2112-11446}, SlimPajama \cite{cerebras2023slimpajama} and C4 \cite{10.5555/3455716.3455856}. We select a subset of these signals and set the thresholds to match the filtering applied for RefinedWeb as closely as possible. A detailed list of the used signals and thresholds can be found in Appendix \ref{appendix:filtering}.
RedPajama-V2 also provides pre-computed LSH signatures which we use to deduplicate the corpus based on a Jaccard distance of 0.8. For a detailed comparison of the quality of RefinedWeb vs RedPajama-V2, we refer to Section \ref{subsec:RW_vs_RP2}. 
For the final data mixture, we sample from RefinedWeb so that we get a full epoch over the course of the whole training.
The rest we fill with data from RedPajama-V2.

In addition to these large-scale corpora, we also include some smaller but presumably high quality datasets.
We use peS2o \cite{peS2o} which consists of scientific papers, stackexchange for Q\&A data, as well as a recent dump from Wikipedia in 5 languages. 
We oversample these clean corpora as described in Section \ref{subsec:additional_datasets}.
We also include some code data from the stack \cite{Kocetkov2022TheStack}. 
The peS2o corpus is used as provided on huggingface datasets \footnote{\url{https://huggingface.co/datasets/allenai/peS2o}}. For "The Stack" we used the same subset as used by StarCoder \cite{li2023starcoder}.

\subsubsection{Machine Translation}
\label{subsubsec:parallel_data}

We include some parallel machine translation data in the training to boost machine translation performance (see Section \ref{subsec:machine_translation}).
We utilize the ParaCrawl corpus \cite{DBLP:conf/acl/BanonCHHHEFKKKO20} for En$\leftrightarrow$\{Es, Fr, De, It\} as well as a smaller in-house corpus from the e-commerce domain.
We serialize the parallel data by concatenating the source and target sentence using a special tag indicating the target language.

\subsection{Tokenizer}
\label{subsec:tokenizer}

Instead of re-using or modifying an existing tokenizer, we decide to train our own tokenizer model on a mix of general and e-commerce domain data spanning 5 languages (English, Spanish, French, German, Italian).
This gives us several advantages, namely (i) full control over the vocabulary including special tokens (ii) better support for multilinguality (iii) better adaptation to e-commerce specific use-cases.

To give an example for the last point, the tokenizer of the LLaMA-2 models splits the word \texttt{'Yugioh'} (the name of a popular trading card game) into 4 separate tokens i.e. \texttt{'\_Y', 'ug', 'io', 'h'}.
Our tokenizer that was trained with e-commerce data, keeps the word as a single token, since it appears frequently in the e-commerce domain.
This means, our LLMs are roughly 4 times faster when generating this word.

We decide to use a vocabulary size of 65k (with byte-level fall back), which is more than twice the size of the LLaMA tokenizer (30k).
A larger vocabulary means that words are split into a smaller number of subwords.
A larger vocabulary is typically beneficial when dealing with multilingual data from very different domains, as we do.

One potential concern regarding larger vocabulary is the increased size of the embedding and projection matrices, which may lead to a slower forward pass in decoding.
We hypothesize that this is not an issue for larger models, since the embedding and projection matrices take up only a very small portion of the overall computations.
We run decoding experiments with the state-of-the-art vLLM framework \cite{kwon2023efficient} with different model and vocabulary sizes.
We find that for model size of 1B, increasing vocabulary size from 30k to 65k leads to an overall slowdown in decoding of less than 5\%.
For model size of 7B, the difference is already less than 1\%.

At the same time, across 10 eBay-internal downstream tasks, we reduce the amount of tokens needed to produce the same output by up to 34\%, resulting in an up to 34\% decoding speedup on these tasks compared to the LLaMA-2 models.

\subsection{Evaluation}
\label{subsec:evaluation}

We evaluate our models in several ways, utilizing public benchmarks in natural language understanding, commonly used machine translation benchmarks, as well as eBay-specific downstream tasks.

\subsubsection{General Domain NLU Tasks}
\label{subsubsec:evaluation_NLU}

We utilize the EleutherAI LM Evaluation Harness \cite{eval-harness} to evaluate the models in terms of general NLU capabilities for English, Spanish, French, German and Italian.
For our preliminary experiments (Section \ref{sec:finding_best_setup}), we follow \cite{DBLP:journals/corr/abs-2311-16867} and use three different aggregates of English tasks in the zero-shot setting: \textbf{zs-1} consists of \texttt{arc\_challenge} \cite{DBLP:journals/corr/abs-1803-05457}, \texttt{arc\_easy} \cite{DBLP:journals/corr/abs-1803-05457}, \texttt{hellaswag} \cite{DBLP:conf/acl/ZellersHBFC19}, \texttt{piqa} \cite{DBLP:conf/aaai/BiskZLGC20} and \texttt{sciq} \cite{DBLP:conf/aclnut/WelblLG17}.
\textbf{zs-2} consists of \texttt{arc\_easy}, \texttt{arc\_challenge}, \texttt{hellaswag}, \texttt{openbookqa} \cite{DBLP:conf/emnlp/MihaylovCKS18}, \texttt{piqa} and \texttt{winogrande} \cite{DBLP:journals/cacm/SakaguchiBBC21}.
\textbf{zs-3} consists of \texttt{hellaswag}, \texttt{lambada} \cite{DBLP:conf/acl/PapernoKLPBPBBF16} and \texttt{piqa}.
These aggregates were found by \cite{DBLP:journals/corr/abs-2311-16867} to have meaningful results and low variability for small model sizes.

For final performance measure of English NLU capabilities we use the same aggregate of zero-shot tasks as \cite{DBLP:journals/corr/abs-2402-00838}, consisting of \texttt{arc\_easy}, \texttt{arc\_challenge}, \texttt{boolq} \cite{DBLP:conf/naacl/ClarkLCK0T19}, \texttt{copa} \cite{DBLP:conf/semeval/GordonKR12}, \texttt{hellaswag}, \texttt{openbookqa}, \texttt{piqa}, \texttt{sciq} and \texttt{winogrande}.

For non-English NLU evaluation, we use \texttt{lambada}, \texttt{xstory\_cloze} \cite{DBLP:journals/corr/abs-2112-10668}, and \texttt{xnli} \cite{DBLP:conf/emnlp/ConneauRLWBSS18} (Spanish), \texttt{lambada}, \texttt{xwinograd} \cite{tikhonov2021heads}, and \texttt{xnli} (French), \texttt{lambada} and \texttt{xnli} (German) and \texttt{lambada} and \texttt{xcopa} \cite{ponti2020xcopa} (Italian).
We purposefully do not include \texttt{mgsm} as it gives very low scores in the zero-shot setting (both for external and internal models) as well as \texttt{pawsx} as it has very high variance in the zero-shot setting.

\subsubsection{Machine Translation}
\label{subsubsec:evaluation_MT}

We report results for commonly used machine translation benchmarks from the WMT Conference on Machine Translation \cite{DBLP:conf/wmt/BarraultBBCFGGH20}.
We use \texttt{newstest09} for En$\leftrightarrow$It, \texttt{newstest13} for En$\leftrightarrow$Es, \texttt{newstest15} for En$\leftrightarrow$Fr and \texttt{newstest20} for En$\leftrightarrow$De and Fr$\leftrightarrow$De.
We report case-sensitive BLEU scores calculated using sacrebleu \cite{DBLP:conf/wmt/Post18}.

\subsubsection{e-commerce-specific Tasks}
\label{subsubsec:evaluation_ebay}

\begin{figure}[h]
    \centering
    \includegraphics[width=0.85\textwidth]{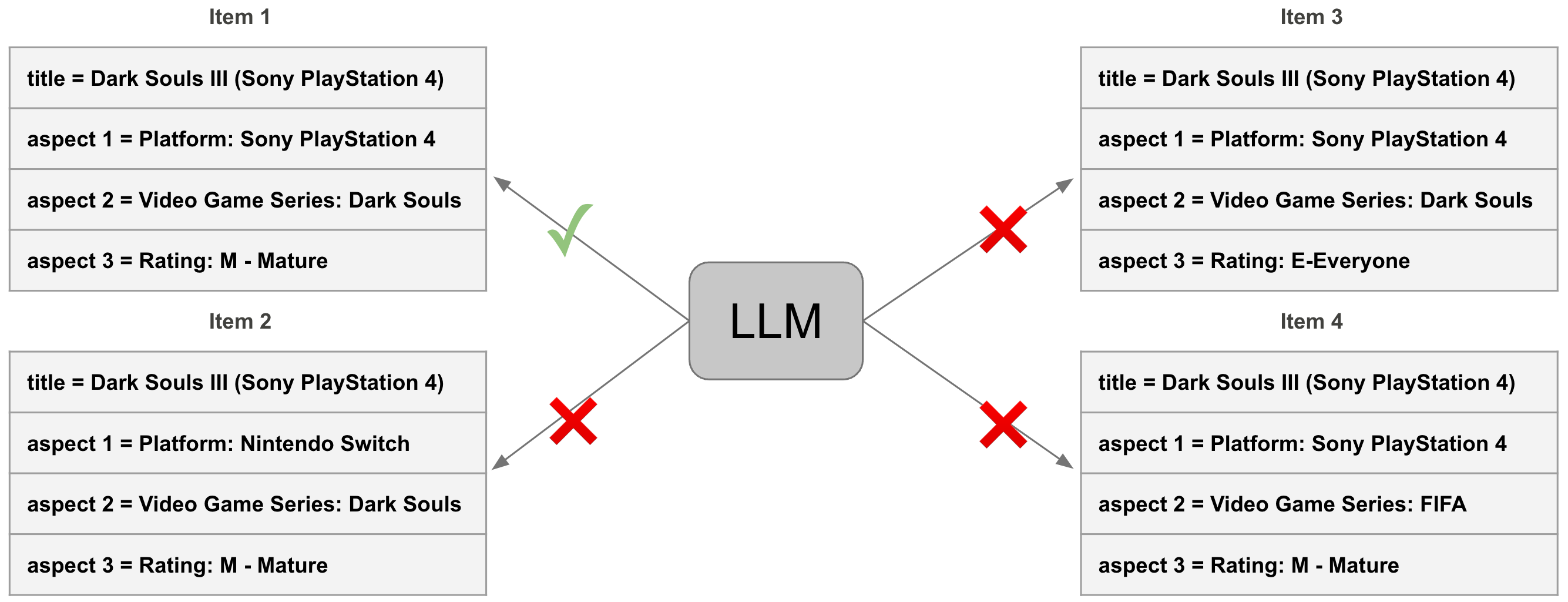}
    \caption{Overview of the item selection task.}
    \label{fig:item_select}
\end{figure}

\begin{figure}[h]
    \centering
    \includegraphics[width=0.85\textwidth]{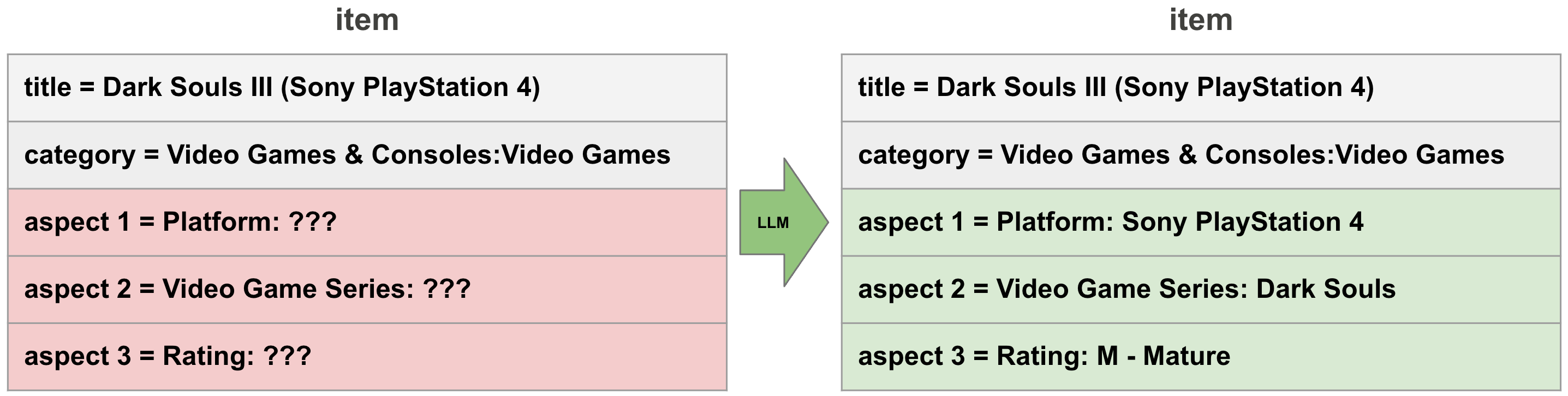}
    \caption{Overview of the aspect prediction task.}
    \label{fig:aspect_predict}
\end{figure}

For evaluation of the model capabilities in the e-commerce setting, we define several tasks. 
\\
\\
1) We calculate perplexity on heldout test sets, consisting of a few million listings on the eBay website.
This gives us a general idea on how well the model is accustomed to the e-commerce domain.
\\
\\
2) We define the \textbf{item selection (IS)} task as shown in Figure \ref{fig:item_select}.
The model has to score 4 item listings (consisting of item title and item aspects), 3 of which were corrupted by replacing some aspects with ones from a similar listing.
If the model gives the best score to the uncorrupted item, it could solve this example successfully.
This task gives us insights in how well the model can associate and link different attributes of an item, in this case title and aspects.
We report accuracy in terms of how many examples are solved correctly.
\\
\\
\newpage
3) We define the \textbf{aspect prediction (AP)} task as shown in Figure \ref{fig:aspect_predict}.
The model is given the title and category of a listing, as well as the corresponding aspect keys.
It has to predict for each aspect key the corresponding aspect value.
For the example in Figure \ref{fig:aspect_predict}, if the aspect key is \texttt{'Rating'}, the model has to predict \texttt{'M-Mature'}, based on the title of the listing.
This task is solving the real problem of automatic aspect prediction for listings, a highly relevant task within eBay.
We compare the predicted aspect values with the ground truth ones and report F1-score.
\section{Finding the Best Setup}
\label{sec:finding_best_setup}

We perform a series of experiments with models of the size of 1B parameters in order to determine the best settings for the large-scale model trainings.
Unless specified otherwise, all models in this section are trained on ca. 500 billion tokens with a batch-size of ca. 4 million tokens.
With the available hardware, these models take just a few hours to train, so we can perform rapid exploration of relevant hyperparameters.
Since we vary the ratio of public vs e-commerce data for different sets of experiments, the numbers between tables are not always comparable.
However, within one table all numbers are always comparable.

\subsection{Comparison to Existing Model}

We start by verifying that our overall data and training setup is working as expected. To do this, we train a 1B parameter model on 500 billion tokens and compare against an open-source model of similar size, namely the Pythia-1B model \cite{DBLP:conf/icml/BidermanSABOHKP23}.
The results are listed in Table \ref{tab:pythia_1b_comparison}.
\begin{table}[h!]
\caption{Comparison against Pythia-1B model.\label{tab:pythia_1b_comparison}}
\centering
\begin{tabular}{l|c|c|c}
Model & zs-1 & zs-2 & zs-3  \\ \midrule
Pythia-1B \cite{DBLP:conf/icml/BidermanSABOHKP23} & 59.37 & 45.06 & 55.72  \\
ours-1B & 61.44 & 47.22 & 56.37  \\
\end{tabular}
\end{table}
We find that our model performance is competitive to existing models and therefore move forward with our experiments.

\subsection{Effect of Randomness}

There are several non-deterministic aspects in the training setup. 
The most important ones are (i) random initialization of the trainable model parameters
(ii) random shuffling of the training data examples.
We aim to quantify the effect of these, in order to determine at which point an improvement on our benchmarks becomes significant in that sense.
We perform 3 training runs with identical settings except for the random seed which effects parameter initialization and data shuffling.
The results are listed in Table \ref{tab:randomness}.
\begin{table}[h!]
\caption{Benchmark results for 3 different training runs with different random seeds.\label{tab:randomness}}
\centering
\begin{tabular}{l|c|c|c}
random seed & zs-1 & zs-2 & zs-3  \\ \midrule
1234 & 56.65 & \textbf{43.93} & 51.35  \\
42 & \textbf{57.38} & 43.08 & 51.37  \\
9999 & 57.34 & 43.59 & \textbf{51.39} 
\end{tabular}
\end{table}
We find that there is some variance on the aggregates we report on.
We conclude that any improvement below 1\,\% absolute is not significant. 

\subsection{Training Hyperparameters}

We try to optimize certain hyperparameters for training.
Specifically we experiment with increasing the variance when randomly initializing model parameters and decreasing the $\beta$ values in the Adam optimizer.
The latter is because we hypothesize that the large batch size we are using means that we need less gradient information from previous update steps.
The results are listed in Table \ref{tab:vary_training_params}.
\begin{table}[h!]
\caption{Benchmark results for different training hyperparameters.\label{tab:vary_training_params}}
\centering
\begin{tabular}{l|l|l|c|c|c}
$\beta_1$ & $\beta_2$ & var. & zs-1 & zs-2 & zs-3  \\ \midrule
0.9 & 0.95 & 0.1 & 56.65 & 43.93 & 51.35  \\
0.8 & 0.9 & 0.1 & \textbf{56.84} & 43.39 & \textbf{51.62}  \\
0.9 & 0.95 & 0.2 & 56.80 & \textbf{44.49} & 51.17 
\end{tabular}
\end{table}
We find that none of the changes improve performance significantly.
We chose $\beta_1 = 0.9$, $\beta_2 = 0.95$, var.$= 0.2$ for the actual training runs.

\subsection{Serializing Structured Data for LLM Training}
\label{subsec:serializing_ebay_data}

As mentioned in Section \ref{subsec:e_commerce}, we experiment with different ways of indicating listing attributes to the model.
We compare using special tags in the vocabulary (like \texttt{'$[$TITLE$]$'}) vs. using tags in natural language (like \texttt{'Item Title:'}) vs. using no indication at all.

The variant of serialization has no significant effect on general domain performance, as can be seen in Table \ref{tab:item_serialization}.
\begin{table}[h!]
\caption{Effect of different listing data serialization strategies on general NLU performance and performance on e-commerce tasks.\label{tab:item_serialization}}
\centering
\begin{tabular}{l|c|c|c|c|c}
serialization & zs-1 & zs-2 & zs-3 & AP & IS  \\ \midrule
tagged & 57.80 & 43.95 & 51.87 & 66.0 & 60.5 \\
natural language tags & \textbf{58.08} & \textbf{44.19} & \textbf{51.93} & \textbf{66.3} & \textbf{68.2}  \\
no tags & 57.01 & 43.17 & 51.60 & 53.6 & 58.5
\end{tabular}
\end{table}
However, we find that the model can best solve the item selection task, if it was trained with natural tagged listing data.
We hypothesise this is because the model can better transfer the knowledge it has learned from the general domain data if the tags are in natural language.
Another argument for using natural tags is that it makes it easier for the model to transfer knowledge from pretraining in the fine-tuning stage where probably no special tags are being used.
In conclusion, we decide to train all our models with the natural tag serialization strategy.

\subsection{Mixing Public-domain and e-commerce Domain Data}
\label{subsec:mixing_public_and_ecommerce}

\begin{figure}[h]
    \centering
    \includegraphics[width=0.55\textwidth]{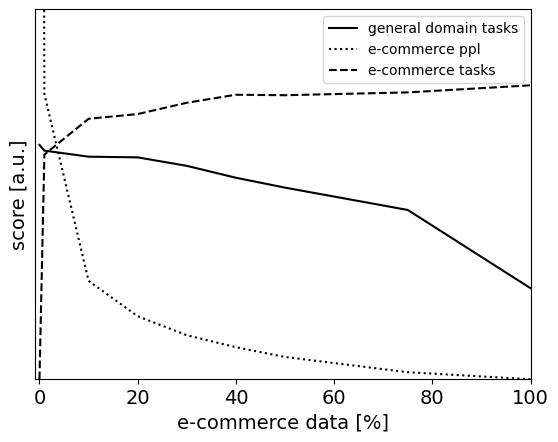}
    \caption{Percentage of e-commerce data in training vs model performance. \lq{}general domain tasks\rq{} indicates the average of \texttt{zs-1}, \texttt{zs-2}, and \texttt{zs-3} (see Section \ref{subsec:evaluation}) and \lq{}e-commerce tasks\rq{} indicates the average of \texttt{aspect prediction} and \texttt{item selection}.}
    \label{fig:ebay_data_percent}
\end{figure}

Another critical question for training is, with which percentage should we mix the general domain and e-commerce domain data. We run a set of experiments, where we vary this ration from 0\% to 100\%, the results of which can be seen in Figure \ref{fig:ebay_data_percent}.
As we include more e-commerce data, we start seeing degradation on the general domain NLU tasks, while at the same time seeing only limited gains on the e-commerce specific tasks. 
We come to the conclusion that a smaller amount of e-commerce data is already enough to enable the model to exceed at e-commerce setting and therefore chose to include 10\% of e-commerce data in the final training data mix.

\subsection{RefinedWeb vs Redpajama-2 Datasets}
\label{subsec:RW_vs_RP2}

As we need large amounts of general domain training data, we investigate different available open source datasets for training.
We train systems on RefinedWeb, as well as on RedPajama-V2.
For the latter, we consider both the unfiltered version as well as after applying filtering and deduplication.
The results can be found in Table \ref{tab:rw_vsrp2}.
\begin{table}[h!]
\caption{RefinedWeb vs RedPajama-V2 datasets.\label{tab:rw_vsrp2}}
\centering
\begin{tabular}{l|c|c|c}
dataset & zs-1 & zs-2 & zs-3  \\ \midrule
RefinedWeb & \textbf{61.49} & \textbf{47.62} & \textbf{57.50} \\
RedPajama-V2 (English-only, last 7 snapshots) & 60.53 & 46.28 & 54.80 \\
\quad + filtering + deduplication (LSH 0.8) & 60.68 & 47.43 & 56.32 
\end{tabular}
\end{table}
We find that, while filtering and deduplication improves the results, RefinedWeb is still slightly higher quality data.
As future work, we plan to improve our filtering and deduplication pipeline, but due to time limitations we could not do this for the current system trainings.
Therefore, for the final trainings of the LiLiuM models, we include RefinedWeb in our training data and fill the rest up with the filtered and deduplicated RedPajama-V2 data.

\subsection{Machine Translation}
\label{subsec:machine_translation}

It has been shown that LLMs have strong capabilities to perform machine translation.
For example, in the latest WMT evaluation for machine translation, GPT4 was among the strongest systems for all tasks translating into English \cite{DBLP:conf/wmt/KocmiABBDFFFGGH23}.
Here, we aim to study the effect of utilizing multilingual and parallel training data on LLM translation capabilities.

We train 1B parameter models using different training data mixtures: (i) just English monolingual data (ii) English monolingual data + non-English monolingual data ($\approx$20\%) (iii) English monolingual data + non-English monolingual data ($\approx$20\%) + parallel data ($\approx$2\%).
To generate translations, we utilize 5-shot decoding for all systems.
For the system trained on parallel data, we also try 0-shot decoding by utilizing the special tags used in training (see Section \ref{subsubsec:parallel_data}).
The results are shown in Table \ref{tab:1b_mt}.
\begin{table}[h!]
\caption{Effect of training data mixture on machine translation performance. We report case-sensitive BLEU scores (see Section \ref{subsubsec:evaluation_MT}).\label{tab:1b_mt}}
\centering
\begin{tabular}{l|c|c|c}
data & En$\xrightarrow{}$X & X$\rightarrow{}$En & De$\leftrightarrow$Fr   \\ \midrule
just English & 3.3 & 14.3 & 1.1  \\
\quad + Multilingual & 17.0 & 24.4 & 10.8  \\
\quad\quad + Parallel &  &  & \\
\quad\quad\quad tagged decoding & \textbf{27.6} & 31.2 & 3.4 \\
\quad\quad\quad 5-shot decoding & 26.0 & \textbf{31.6} & \textbf{17.5}
\end{tabular}
\end{table}
In general, the LLMs perform better when translating into English vs into a different language.
Adding non-English monolingual data already boosts the translation performance significantly.
Adding parallel data further improves translation quality by a significant margin, even though the data size is relatively small ($\approx$2\%).
Also, the De$\leftrightarrow$Fr translation quality improves even though we do not add any explicit parallel data for that language pair.

Another interesting observation is that 0-shot decoding with tags and 5-shot decoding without tags perform very similar.
This indicates that the model is quite able to generalize from the parallel data, even though it is preprocessed in a very specific way using special tags.
Although this generalization does not always work, as can be seen for De$\leftrightarrow$Fr. 

\subsection{High Quality Corpora}
\label{subsec:additional_datasets}

As discussed in Section \ref{subsubsec:data_general_domain}, we also include smaller, high quality datasets into our pretraining data mix.
As can be seen from Table \ref{tab:add_datasets}, including these datasets slightly improves model performance.
\begin{table}[h!]
\caption{Effect of additional datasets on model quality.\label{tab:add_datasets}}
\centering
\begin{tabular}{l|c|c|c}
data & zs-1 & zs-2 & zs-3  \\ \midrule
just RedPajama-V2 & 59.67 & 45.70 & 54.26 \\
\quad + high quality & 60.17 & 46.59 & 54.45 \\
\quad\quad + oversample high quality & \textbf{60.53} & \textbf{46.71} & \textbf{54.69} \\
\quad + code data & 59.64 & 46.06 & 54.41 \\
\end{tabular}
\end{table}
Oversampling the high quality data leads to further minor improvements.
In the final training setups, we oversample the small datasets so that they are being seen twice over the course of 3 trillion tokens.
We also verify that adding a small percentage of code data does not hurt general domain performance of the model.
In the final training data we decide to sample 5\% of our data from code in order to improve code understanding and reasoning capabilities.

\subsection{The Best 1B model Architecture}
\label{subsec:best_1b_arch}

So far we have used the same architecture for all of the 1B parameter models that we have trained. 
We hypothesize that adding more layers with less parameters per layer could lead to further improvements.
Training models with different architectures leads to the results shown in Table \ref{tab:best_1B_arch}.
\begin{table}[h!]
\caption{Different architectures for the 1B model.\label{tab:best_1B_arch}}
\centering
\begin{tabular}{l|l|l|l|l|c|c|c}
\#params & \#layers & hidden-size & ffn-hidden-size & \#attn. heads & zs-1 & zs-2 & zs-3  \\ \midrule
1.14B & 16 & 2048 & 6,144 & 16 &  61.27 & 47.56 & 56.36 \\
1.15B & 22 & 1792 & 5,376 & 16 &  61.21 & 47.51 & 56.57 \\
1.27B & 24 & 2048 & 4,096 & 16 &  \textbf{61.78} & 48.62 & 57.49 \\
1.27B & 24 & 2048 & 4,096 & 32 &  61.18 & 48.17 & 56.32 \\
1.40B & 22 & 2048 & 5,632 & 16 &  61.76 & \textbf{48.79} & \textbf{57.59} \\
\end{tabular}
\end{table}
In general, more layers help but more attention heads per layer does not.
We decide to use the setting from the third row for our 1B model, since it seems the best compromise for performance vs number of parameters. 

\newpage

\section{LiLiuM Models}
\label{sec:lilium_models}

In this section, we describe the final system trainings and evaluations for the LiLiuM models.

\subsection{Training}

For the LiLiuM 1B, 7B and 13B models, we use the setups described in Table \ref{tab:lilium_setup}.
\begin{table}[h!]
\caption{Overview of the LiLiuM models.\label{tab:lilium_setup}}
\centering
\def\arraystretch{1.2}
\begin{tabular}{l|ccc} 
 & \textbf{LiLiuM 1B} & \textbf{LiLiuM 7B} & \textbf{LiLiuM 13B}  \\ \midrule
\textbf{data} & & & \\ 
\# tokens & 3,000B & 3,000B & 3,000B \\
e-commerce & 10\% & 10\% & 10\% \\ 
code & 5\% & 5\% & 5\% \\ 
non-English & 20\% & 20\% & 20\% \\ 
translation & 1-2\% & 1-2\% & 1-2\% \\ \midrule
\textbf{architecture} & & & \\
\# layers & 24 & 32 & 40 \\ 
\# heads & 16 & 32 & 40 \\ 
hidden-size & 2,048 & 4,096 & 5,120 \\
ffn-hidden-size & 4,096 & 11,008 & 13,824 \\
context-size & 4,096 & 4,096 & 4,096 \\
vocab-size & 65,024 & 65,024 & 65,024 \\ \midrule
\textbf{training} & & & \\
batch-size (tokens) & 6.6M & 6.6M & 6.6M \\
lr & 3.0e-4 & 3.0e-4 & 3.0e-4 \\
min-lr & 3.0e-5 & 3.0e-5 & 3.0e-5 \\
TP & 1 & 1 & 2 \\
PP & 1 & 2 & 4 \\
DP & 800 & 400 & 100 \\
time (GPU hours) & 72k & 307k & 538k \\
\end{tabular}
\end{table}

For the hyperparameters we mostly follow existing works.
We use the Adam optimizer \cite{DBLP:journals/corr/KingmaB14} in a mixed-precision setting (bf16) with weight decay 0.1 and a cosine learning rate schedule across all update steps.
We chose a larger batch size, since our data is quite diverse in terms of languages and domain.

Model training went mostly smooth without major issues.
Like other groups \cite{DBLP:journals/corr/abs-2211-05100, DBLP:journals/corr/abs-2311-16867, young2024yi}, we experienced several crashes due to hardware related issues like connection timeouts or
 hardware failures.
These can hardly be avoided when scaling up the hardware, and we minimize the impact of these crashes by frequently writing checkpoints (every 1-2 hours) and automatically restarting the training job in case of a crash.

\begin{figure}[h]
    \centering
    \includegraphics[width=0.575\textwidth]{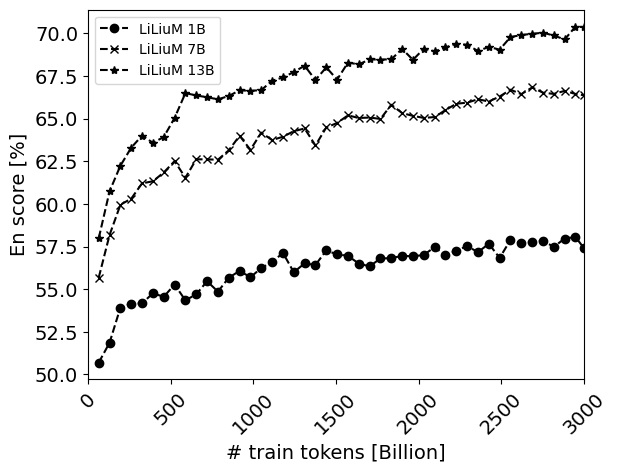}
    \caption{Performance on the downstream English NLU task aggregate (see Section \ref{subsubsec:evaluation_NLU}) during training.}
    \label{fig:chkpt_training_perf}
\end{figure}

It is important to monitor training speed in addition to training and validation loss, since on one occasion we experienced a slowdown (but no crash) in training due to an overheating hardware component.
During training, we also regularly evaluate the model checkpoints in terms of downstream performance to ensure that the training is on the right track to reach the expected final model performance (see Figure \ref{fig:chkpt_training_perf}).
We find that model performance continuously increases over the course of the training.

During the training of the 13B model, we experienced several loss spikes also reported by other groups \cite{DBLP:journals/jmlr/ChowdheryNDBMRBCSGSSTMRBTSPRDHPBAI23, DBLP:journals/corr/abs-2311-16867}.
In all cases, the training could be resumed without manual intervention, most likely thanks to gradient clipping and bf16 mixed precision.

\subsection{Checkpoint Averaging Improves LLM Performance}

One question for any neural network training is, how to select the final model checkpoint after the training is finished.

\begin{figure}[h]
    \centering
    \includegraphics[width=0.55\textwidth]{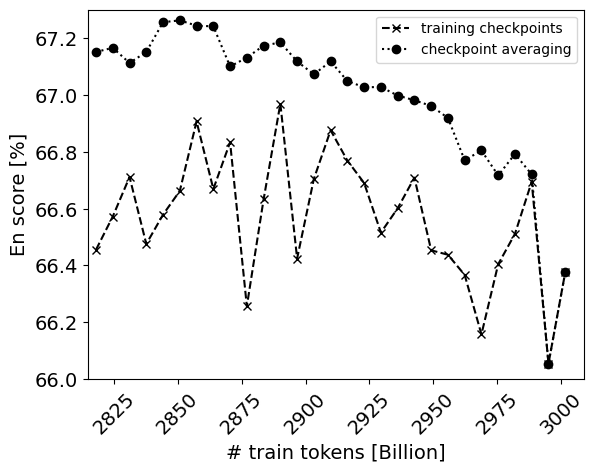}
    \caption{Performance of LiLiuM 7B model checkpoints on the downstream English NLU task aggregate (see Section \ref{subsubsec:evaluation_NLU}). Checkpoint averaging indicates averaging the model parameters of all checkpoints between the current and the last checkpoint.}
    \label{fig:chkpt_avg}
\end{figure}

The simplest way is to just select the last checkpoint of the training, as this does not require storing any intermediate checkpoints.
However, since the model is in the final stages of training, the last checkpoint might not necessarily be the best one for a given set of tasks.
For example in Figure \ref{fig:chkpt_avg}, we plot the performance of the last few checkpoints of the LiLiuM 7B model training on the aggregate of the English NLU tasks (dashed line).
Clearly, the last checkpoint is not the strongest one on this set of tasks.

A common method to select the best checkpoint is called \lq{}early stopping\rq{}: we evaluate all checkpoint in terms of validation set perplexity and select the best one this way.
This can be helpful to avoid overfitting on the training data.
In case of the LLM training, we typically train the model for only a single epoch, therefore overfitting is less of an issue.
We verify this by checking the validation perplexity during training. It consistently decreases and the checkpoint with the best validation perplexity is one of the last ones created.
However, we also find that validation perplexity is not very well correlated with our downstream performance measures towards the end of the training.

There is the option to select the best checkpoint according to downstream task performance, e.g. score on English NLU tasks.
However, this is bad scientific practice since we are overfitting on the test set in this case.
Also, it means that most likely the checkpoint is not better on other test sets that were not part of the selection process.
We can prove this hypothesis by selecting the best checkpoint of the LiLiuM 7B model for the English NLU benchmarks and then evaluating this checkpoint on the benchmarks for the other languages.
The results can be seen in Table \ref{tab:chkpt_avg}.
The best checkpoint for the English NLU benchmarks \lq{}\texttt{best checkpoint (En)}\rq{} does not perform well for most other languages and actually performs slightly worse than the last checkpoint of the training on average.

\begin{table}[h!]
\caption{LiLiuM 7B: Performance of different model checkpoints on NLU tasks. We report on the language specific aggregates defined in Section \ref{subsubsec:evaluation_NLU}. avg is the average over all languages.\label{tab:chkpt_avg}}
\centering
\def\arraystretch{1.2}
\begin{tabular}{l|ccccc|c}
model- & \multicolumn{6}{c}{NLU} \\ 
checkpoint & En & Es & Fr & De & It & avg \\ \midrule
last checkpoint & 66.4 & 53.3 & 59.0 & 44.9 & 60.3 & 56.8 \\
best checkpoint (En) & 67.0 & 53.1 & 58.6 & 45.1 & 59.9 & 56.7 \\
cont. train \cite{DBLP:journals/corr/abs-2402-00838} & 66.9 & \textbf{53.5} & \textbf{60.0} & 45.5 & 61.0 & \textbf{57.4} \\
average last 20 checkpoints & \textbf{67.3} & \textbf{53.5} & 59.2 & \textbf{45.8} & \textbf{61.3} & \textbf{57.4} \\
\end{tabular}
\end{table}

We propose to instead use checkpoint averaging to obtain the final model checkpoint.
Checkpoint averaging has shown to be effective for related tasks such as machine translation \cite{DBLP:conf/nips/VaswaniSPUJGKP17, DBLP:conf/ijcnlp/GaoHYN22a} and automatic speech recognition \cite{DBLP:conf/icassp/DongXX18, DBLP:conf/asru/KaritaWWYZCHHIJ19}.
It has also been shown that checkpoint averaging can help LLM training convergence \cite{sanyal2023early} and cross-lingual knowledge transfer \cite{DBLP:conf/acl/SchmidtVG23}.
For any given checkpoint, we calculate the checkpoint average by averaging parameters from all checkpoints between the selected one and the last checkpoint.
This results in the list of checkpoints depicted in Figure \ref{fig:chkpt_avg} (dotted line).
We find that averaging checkpoints from the last $\approx$20k iterations results in the best performance, although this parameter most likely depends on the checkpoint frequency, learning rate, etc.

Recently, \cite{DBLP:journals/corr/abs-2402-00838} suggest to continue training the final model checkpoint for an additional 1,000 steps while reducing the learning rate linearly to 0.
We also test this approach (\texttt{cont.\,train} in Table \ref{tab:chkpt_avg}) and find that this technique also improves performance consistently across tasks.
However, it has the disadvantage that it requires large amounts of compute for the continued training while checkpoint averaging does not.

We conclude that checkpoint averaging is the most robust way of selecting the best model checkpoint and, as additional benefit, it comes at no additional training cost. 
We perform checkpoint averaging for all three LiLiuM models and find that using checkpoints from the last $\approx$20k iterations works best for us.

\subsection{Evaluation}

In Table \ref{tab:lilium_final_performance}, we report the performance of the LiLiuM 1B, 7B and 13B models in comparison to LLaMA-2.
\begin{table}[h!]
\caption{Final model performance on different downstream tasks. For NLU, we report on the language specific aggregates defined in Section \ref{subsubsec:evaluation_NLU}. For MT, we report the average BLEU score on the test sets defined in Section \ref{subsubsec:evaluation_MT}. For e-commerce, we report on the tasks we defined in Section \ref{subsubsec:evaluation_ebay}. \label{tab:lilium_final_performance}}
\centering
\def\arraystretch{1.2}
\begin{tabular}{l|ccccc|cc|cc}
\multirow{2}{*}{model} & \multicolumn{5}{c|}{NLU} & \multicolumn{2}{c|}{MT} & \multicolumn{2}{c}{e-commerce} \\ 
 & En & Es & Fr & De & It & X$\xrightarrow{}$En & En$\xrightarrow{}$X & AP & IS \\ \midrule
\textbf{1B} &  &  &  &  & & & & & \\
LiLiuM 1B & \textbf{58.4} & \textbf{48.7} & \textbf{53.9} & \textbf{41.9} & \textbf{50.4} & \textbf{32.9} & \textbf{27.4} & \textbf{49.1} & \textbf{69.0} \\
Pythia-1B \cite{DBLP:conf/icml/BidermanSABOHKP23} & 54.3 & 40.1 & 43.9 & 33.3 & 40.4 & 21.4 & 13.8 & 22.3 & 53.7 \\ \midrule
\textbf{7B} &  &  &  &  & & & & & \\
LiLiuM 7B & 67.3 & \textbf{53.5} & \textbf{59.2} & \textbf{45.8} & \textbf{61.3} & \textbf{38.4} & \textbf{34.5} & \textbf{63.8} & \textbf{75.0} \\
LLaMA-2 7B \cite{DBLP:journals/corr/abs-2307-09288} & \textbf{68.1} & 51.0 & 56.1 & 43.1 & 56.2 & 36.2 & 27.8 & 32.3 & 36.7 \\ \midrule
\textbf{13B} &  &  &  &  & & & & & \\
LiLiuM 13B & 70.8 & \textbf{57.2} & \textbf{59.7} & \textbf{48.4} & \textbf{66.0} & \textbf{39.4} & \textbf{36.2} & \textbf{55.6} & \textbf{75.3} \\
LLaMA-2 13B \cite{DBLP:journals/corr/abs-2307-09288} & \textbf{71.0} & 53.3 & 58.5 & 45.7 & 59.1 & 37.8 & 30.8 & 34.8 & 21.2 \\ 
\end{tabular}
\end{table}
For the AP task, we use 1-shot decoding for all models to have a more fair comparison to external models, which have not seen this data format in training.

We find that the LiLiuM models perform similar to the respective LLaMA-2 models on English NLU tasks.
The small gap to LLaMA-2 is most likely because we include significant amounts of non-English and e-commerce specific data in our pretraining.
This may lead to some performance loss on English NLU benchmarks as we discuss in Section \ref{subsec:mixing_public_and_ecommerce}.
On non-English NLU tasks, we outperform the respective LLaMA-2 model by 3.5 points on average.
In fact, our smaller 7B model already outperforms the larger LLaMA-2 13B.
While LLaMA-2 was trained on English-centric data only, we find it quite impressive that adding a relatively small amount of non-English data ($\sim$5\% per language) leads to such large improvements.
LiLiuM also outperforms LLaMA-2 on machine translation tasks.

Regarding the the e-commerce specific downstream tasks, the LiLiuM models outperform LLaMA-2 by a large margin.
This highlights the deep knowledge of the LiLiuM models about the e-commerce domain.
Currently we are evaluating the models on more concrete e-commerce use-cases, where they also show promising improvements over general domain, third-party models.
 
\section{Conclusion and Future Work}

In this paper, we discuss the development of the LiLiuM family of large language models with up to 13 billion parameters.
These models were build from scratch by eBay, including the training of our own tokenizer, customized towards the e-commerce domain. 
LiLiuM 1B/7B/13B were each trained on 3 trillion tokens of multilingual text data from general and e-commerce domain.
This demonstrates eBay's capabilities to efficiently train such models. 

On English NLU benchmarks, the LiLiuM models perform similar to the LLaMA-2 models.
On non-English NLU tasks, machine translation, and on e-commerce specific downstream tasks, the LiLiuM models outperform LLaMA-2.

We show that, by customizing the model vocabulary towards the e-commerce domain, we can significantly increase decoding speed for eBay-specific use-cases.
Furthermore, we share our insights in using the newly released RedPajama-V2 dataset for LLM pretraining.
We also discuss how to best serialize structured data for autoregressive model training.
Finally, we show that adding a small amount of parallel data into pretraining significantly boosts translation performance.

We compare different strategies of extracting the final model checkpoint after training and find checkpoint averaging to be the most robust method.

Moving forward, we are putting our focus on improving specific aspects of our model development pipeline.
Specifically we
\begin{enumerate}
    \item continue improving our data pipeline, in regards to filtering, deduplication and PII removal.
    \item are working on including further types of eBay-specific data in the model training.
    \item train larger models to improve model quality.
    \item utilize the Mixture-of-Experts architecture \cite{DBLP:conf/iclr/ShazeerMMDLHD17, DBLP:journals/jmlr/FedusZS22} to improve training and inference efficiency.
\end{enumerate}

Also, we will move to our own in-house GPU cluster for future system trainings.

\bibliographystyle{unsrt}  
\bibliography{sample}  

\newpage
\appendix
\section{RedPajama-V2 Filtering}
\label{appendix:filtering}
\begin{table}[h!]
\caption{Quality signals and thresholds used for filtering of the RedPajama-V2 dataset.}
\small
\centering
\begin{tabular}{l|p{0.40\linewidth}|p{0.09\linewidth}|p{0.10\linewidth}}
quality signal & description & used in & threshold \\ \hline
ccnet\_language\_score &
  score of the language identification model &
  CCNet &
  $\textgreater 0.65$ \\
ccnet\_length &
  number of characters &
  CCNet &
  $\textgreater 200$ \\
rps\_doc\_frac\_lines\_end\_with\_ellipsis &
  The fraction of lines that end with an ellipsis, where an ellipsis is defined as either "..." or "…". &
  RefinedWeb, Gopher &
  $\textless 0.3$ \\
rps\_doc\_frac\_no\_alph\_words &
  The fraction of words that contain no alphabetical character. &
  RefinedWeb, Gopher &
  $\textless 0.2$ \\
rps\_doc\_lorem\_ipsum &
  The ratio between the number of occurrences of 'lorem ipsum' and the number of characters in the content after normalisation. &
  C4 &
  $== 0$ \\
rps\_doc\_mean\_word\_length &
  The mean length of words in the content after normalisation &
  RefinedWeb, Gopher &
  $\textgreater{}3$ and \newline $\textless 10$ \\
rps\_doc\_stop\_word\_fraction &
  The ratio between the number of stop words and the number of words in the document. Stop words are obtained from the stopwords-json repo. &
  RefinedWeb, Gopher &
  $\textgreater 0$ \\
rps\_doc\_symbol\_to\_word\_ratio &
  The ratio of symbols to words in the content.. Symbols are defined "\#", "...", and "…". &
  RefinedWeb, Gopher &
  $\textless 0.1$ \\
rps\_doc\_word\_count &
  The number of words in the content after normalisation. &
  RefinedWeb, Gopher &
  $\textgreater{}50$ and \newline $\textless 100000$ \\
rps\_lines\_start\_with\_bulletpoint &
  Whether the lines that start with a bullet point symbol. The following set of unicodes are considered a bullet point: \textbackslash{}u2022 (bullet point), \textbackslash{}u2023 (triangular bullet point), \textbackslash{}u25B6 (black right pointing triangle), \textbackslash{}u25C0 (black left pointing triangle), \textbackslash{}u25E6 (white bullet point), \textbackslash{}u25A0 (black square), \textbackslash{}u25A1 (white square), \textbackslash{}u25AA (black small square), \textbackslash{}u25AB (white small square), \textbackslash{}u2013 (en dash). &
  RefinedWeb, Gopher &
  $ratio \textless 0.9$ \\
rps\_doc\_frac\_chars\_dupe\_5grams &
  The fraction of characters in duplicate word 5grams &
  RefinedWeb, Gopher &
  $\textless 0.15$ \\
rps\_doc\_frac\_chars\_dupe\_6grams &
  The fraction of characters in duplicate word 6grams &
  RefinedWeb, Gopher &
  $\textless 0.14$ \\
rps\_doc\_frac\_chars\_dupe\_7grams &
  The fraction of characters in duplicate word 7grams &
  RefinedWeb, Gopher &
  $\textless 0.13$ \\
rps\_doc\_frac\_chars\_dupe\_8grams &
  The fraction of characters in duplicate word 8grams &
  RefinedWeb, Gopher &
  $\textless 0.12$ \\
rps\_doc\_frac\_chars\_dupe\_9grams &
  The fraction of characters in duplicate word 9grams &
  RefinedWeb, Gopher &
  $\textless 0.11$ \\
rps\_doc\_frac\_chars\_dupe\_10grams &
  The fraction of characters in duplicate word 10grams &
  RefinedWeb, Gopher &
  $\textless 0.10$ \\
rps\_doc\_frac\_chars\_top\_2gram &
  The fraction of characters in the top word 2gram. &
  RefinedWeb, Gopher &
  $\textless 0.20$ \\
rps\_doc\_frac\_chars\_top\_3gram &
  The fraction of characters in the top word 3gram. &
  RefinedWeb, Gopher &
  $\textless 0.18$ \\
rps\_doc\_frac\_chars\_top\_4gram &
  The fraction of characters in the top word 4gram. &
  RefinedWeb, Gopher &
  $\textless 0.16$ \\
rps\_doc\_ldnoobw\_words &
  The number of sequences of words that are contained in the List-of-Dirty-Naughty-Obscene-and-Otherwise-Bad-Words blocklist. The blocklist is obtained from the LDNOOBW repo. &
  C4 &
  $\textless 5$ \\
rps\_doc\_ut1\_blacklist &
  A categorical id corresponding to the list of categories of the domain of the document. Categories are obtained from the UT1 blacklist. The list is obtained from UT-Capitole. &
  RefinedWeb &
  URL domain not in blacklist \\
\end{tabular}
\end{table}

\end{document}